\documentclass[twoside]{article}
\usepackage{ukai}

\usepackage{booktabs}
\usepackage{adjustbox}
\usepackage{float}
\usepackage{cleveref}

\crefname{appendix}{appendix}{appendices}
\Crefname{appendix}{Appendix}{Appendices}
\renewcommand{\confyear}{2026}

\title{\customtitle{An AI-Based Decision-Support Pipeline for Day-Ahead Photovoltaic Forecasting\thanks{This paper has been accepted for publication in the Proceedings of the UK AI Conference (UK-AI 2026).}}}

\setauthorsshort{Dehghan et al.}

\author{
    Fariba Dehghan, Sebastian Stein, Vahid Yazdanpanah, Stephanie Gauthier, Masood Nazari\\
    {University of Southampton}\\
    \small
    \begin{tabular}{c}
        \texttt{F.Dehghan@soton.ac.uk, ss2@ecs.soton.ac.uk, V.Yazdanpanah@soton.ac.uk}\\
        \texttt{S.Gauthier@soton.ac.uk, M.Nazari@soton.ac.uk}
    \end{tabular}
}

\date{}

\begin{document}
\maketitle

\begin{abstract}
Reliable photovoltaic (PV) forecasts can support low-carbon energy systems, but deployed sites may have only short and incomplete records. Physical and hybrid methods can be sensitive to weather inputs, calibration, and timestamp-alignment, while individual machine learning models may capture different parts of the forecasting problem. We study hourly day-ahead PV forecasting at a United Kingdom charging station using one year of inverter measurements, with $9.25\%$ of hours missing. The pipeline checks timestamp-alignment, derives solar and clearness features, adds short-term weather context, and combines five complementary models using non-negative least squares stacking, with the combination fitted only on validation observations. We compare against smart persistence, a weather-scaled baseline that carries the previous day's PV behaviour forward using target-day irradiance. With retrospective weather, the combined model reduces daylight normalised root mean square error (RMSE) by $31.2\%$ under random day-fold evaluation and by $2.9\%$ under rolling-origin evaluation, although the latter improvement is not robust across days. It also improves by $3.0\%$ over the single model selected from validation performance. Replacing retrospective weather with a public product sampled at a constant 24-hour lead increases daylight RMSE by $13.1\%$ and $4.2\%$ under the two protocols, while retaining positive skill over smart persistence.
\end{abstract}


\section{Introduction}
\label{sec:introduction}

Machine learning methods are widely used for forecasting solar and environmental quantities~\cite{antonanzas2016review,voyant2017machine}. We study hourly day-ahead photovoltaic (PV) forecasting at the Future Electric Vehicle Energy Networks supporting Renewables (FEVER) demonstrator at Wide Lane, Southampton, United Kingdom, a container-based off-grid charging testbed in which PV generation is an input to charging and storage decisions~\cite{fever_project}. Day-ahead here means predicting the next 24 hours at hourly resolution, which gives a full next-day generation profile before the target day begins. At this horizon weather can be used as an input, while changes in cloud conditions remain a main source of forecast uncertainty~\cite{antonanzas2016review,voyant2017machine}. PV generation is also bounded, seasonal, and strongly affected by weather, so both the representation of physical conditions and the evaluation design matter. Cross-validation that ignores time order can be valid for autoregressive forecasting under stated assumptions about the errors~\cite{bergmeir2018note}; energy forecasting typically uses chronological or rolling evaluation when the aim is later use~\cite{hong2016probabilistic,yang2019ropes}. Uncertainty in the available PV energy can affect how charging and storage resources are scheduled, which is why the reported forecast error is relevant at this site.

Existing methods represent the problem in different ways. Physical and hybrid approaches use solar and weather information, statistical methods exploit temporal dependence, and machine learning models can learn nonlinear relationships from multivariate inputs~\cite{lorenz2009irradiance,bacher2009online,friedman2001greedy,lecun1998gradient}. These approaches therefore emphasise different structures in the same record: the solar generation scale, short-term persistence, and nonlinear weather--power relationships, which is why we evaluate how those representations work together rather than a model architecture alone. We therefore ask four empirical questions. First, how does timestamp-alignment between site measurements and gridded weather inputs affect physical estimates and learned models? Second, how do direct solar and clearness input features affect performance when a plane-of-array (POA)-normalised representation is already present? Third, can models with different error patterns be combined to improve forecast accuracy? Fourth, how much does measured performance change when retrospective weather data are replaced by a public forecast product taken at a constant lead of 24 hours?

\subsection{Contributions}
\label{sec:contributions}

This paper makes the following contributions:

\begin{itemize}

\item \textbf{An explicit timestamp-alignment diagnostic.}
We identify an apparent one-hour discrepancy between PV output and gridded irradiance and evaluate candidate corrections using a parameter-free physical diagnostic and training-only fold checks. A one-hour backward shift raises the daylight coefficient of determination ($R^2$) of the irradiance-based alignment measure from $0.42$ to $0.60$, although the learned forecasting models perform slightly better without the shift. We therefore retain the correction as an empirically supported timestamp convention while reporting its forecasting cost explicitly.

\item \textbf{A leakage-safe physical representation and ablation.}
We construct solar geometry, irradiance, and POA-normalised representations without using held-out target information and explicitly test their value. The benefit is protocol-dependent: short-term weather context helps rolling-origin forecasting, whereas dropping the raw solar and clearness inputs while keeping the POA-normalised target lowers rolling-origin error. We report these diagnostics rather than changing the headline pipeline.

\item \textbf{A validation-fitted ensemble.}
We combine five models with different structure. Non-negative least squares (NNLS) stacking is the pre-specified reported combination, because non-negativity avoids cancellation between learners and makes the fitted combination easier to interpret, and it is the ensemble reported throughout the headline, ablation, and weather-input results. It reduces daylight root mean square error (RMSE) relative to the validation-selected base learner by $6.7\%$ under random day-fold evaluation and $3.0\%$ under rolling-origin evaluation, with conditional day-level bootstrap intervals that exclude zero under both protocols. The remaining combination methods are reported as a comparison only; identifying the lowest-error one requires the test metrics.
\item \textbf{A sensitivity analysis of forecast weather inputs.}
The main experiments use retrospective weather, which is more informative than the forecasts available in practice. We therefore repeat the pipeline using a public weather product in which each hourly value was predicted 24 hours earlier, and compare it with a retrospective control using the same weather variables. Daylight RMSE increases by $13.1\%$ under random day-fold evaluation and by $4.2\%$ under rolling-origin evaluation, while skill relative to smart persistence remains positive. Because the 24-hour-lead profile is assembled from successive weather updates rather than one forecast issuance, this is a sensitivity test rather than a full operational replay.

\end{itemize}

All code and scripts needed to reproduce the analysis, subject to access to the site measurements, are available in the project repository\footnote{\url{https://github.com/FdehghanSoton/AI_Pipeline_PV_Forecasting}}.
\Cref{sec:related} reviews related work, \Cref{sec:data,sec:models} describe the data and forecasting pipeline, \Cref{sec:evaluation} presents the evaluation and results, and \Cref{sec:conclusion} summarises the findings and limitations.

\section{Related Work}
\label{sec:related}

Guidance for energy forecasting emphasises reproducible methods and evaluation procedures that match the intended setting~\cite{hong2016probabilistic,yang2019ropes}. Physical and hybrid approaches often use weather information together with irradiance and PV system models to estimate power output~\cite{lorenz2009irradiance}. Using those products with site measurements requires care because the sources can differ in spatial scale, bias, and timestamp convention. Authors in~\cite{lorenz2009irradiance} correct bias in forecast irradiance using ground observations before converting the irradiance to PV power. Reference~\cite{yang2018nsrdb} shows that errors reported in a validation of a satellite-derived irradiance database were affected by how the two series were aggregated, and that correcting the temporal treatment substantially changed the reported error. Hourly records may also describe intervals using different timestamp conventions, so matching timestamps need not represent the same interval. Solar Data Tools includes automatic timestamp cleaning and time-shift detection, including shifts associated with daylight saving~\cite{meyers2020solardatatools}. Authors in~\cite{perry2022shift} document a different recording problem: abrupt shifts in PV power and irradiance caused by hardware replacement or software changes rather than a change in the plant. These studies are why we check temporal and recording consistency before fitting.

Forecasting models use the same temporal and weather information in different ways. Authors in~\cite{bacher2009online} combine autoregressive structure with numerical weather predictions and use a clear-sky-based transform so the model does not have to relearn the daily and seasonal solar pattern from the power series. Gradient boosting and convolutional neural networks (CNNs) are standard other representations of tabular and grid-structured inputs~\cite{friedman2001greedy,lecun1998gradient}. Reviews of solar forecasting describe these machine learning approaches alongside statistical and physical methods~\cite{antonanzas2016review,voyant2017machine}. Authors in~\cite{bates1969combination} show that a forecast combination depends on both the errors of the members and the dependence between those errors; stacked generalisation learns the combination from the members' predictions~\cite{wolpert1992stacked}. Published guidance asks for evaluation against stated reference methods~\cite{yang2019ropes,yang2019standard}. These results are why we use structurally different base learners, learn combination weights from validation predictions, and score the pipeline against the references in \Cref{sec:metrics}.

\section{Data and Preprocessing}
\label{sec:data}

This section describes the data used in the forecasting pipeline. We first describe the site measurements and their treatment in \Cref{sec:site-data},
then the weather inputs in \Cref{sec:weather-inputs}, and finally the solar geometry and short-term temporal features in
\Cref{sec:solar-features}.

\subsection{Site and PV Measurements}
\label{sec:site-data}

The case study uses an hourly PV active power series recorded at the container-based FEVER demonstrator at Wide Lane, Southampton, United Kingdom, at assumed coordinates $50.91^{\circ}$N, $1.40^{\circ}$W and an altitude of $30$~m. We treat FEVER as an off-grid charging station setting~\cite{fever_project}, where PV generation is an input to charging and storage decisions. The array is assumed to have a fixed tilt of $30^{\circ}$ and a true-south ($180^{\circ}$) azimuth; these are approximate assumptions, not surveyed parameters. The prediction target $Y_\tau^{\mathrm{PV}}$ is the inverter channel (solar power in watts) at hourly timestamp $\tau$.

The raw record covers 5 March 2025 to 1 April 2026 and is reindexed onto a regular Coordinated Universal Time (UTC) hourly grid. The missing hour pattern is consistent with local-time recording: the record spans the British Summer Time spring transitions on 30 March 2025 and 29 March 2026, and the corresponding 01:00~UTC hour is absent on both dates. These are two of only three isolated one-hour gaps; the remaining missing hours are concentrated in six longer gaps of at least ten hours. A local-clock logger would omit the skipped spring-forward hour, but this does not by itself identify the timestamp convention, so we test candidate corrections directly in \Cref{sec:ablation}.
Negative PV values are clipped to zero. Missing measurements receive a placeholder value of zero and a missingness indicator
$\mathbf{1}\{\tau\in\mathcal{M}\}$, where $\mathcal{M}$ denotes unobserved hours; these placeholders are never treated as measured zeros. The CNN availability channel is the complement $a_\tau=\mathbf{1}\{\tau\notin\mathcal{M}\}$, which equals one on observed hours. In total, 868 hours are missing ($9.25\%$ of the hourly grid). The causes of the longer gaps are not recorded and may reflect interruptions in measurement or data acquisition. Tabular models train only on observed target hours, and all reported metrics exclude $\mathcal{M}$. For the CNN in \Cref{sec:models}, $a_\tau$ is also supplied as an input channel, while missing target hours receive zero loss weight, so a complete 24-slot calendar day need not contain 24 observed measurements. We define the empirical plant capacity as $ C^{\mathrm{PV}}= Q_{0.999}\left(Y_\tau^{\mathrm{PV}}\right),$
where $Q_{0.999}$ is the 99.9th percentile of observed PV power. This reduces the influence of isolated extreme values and gives $C^{\mathrm{PV}}\approx1664$~W ($\approx1.66$~kW) for the full record. To avoid using held-out targets, the POA-normalised target, the CNN normalisation, and the reference forecasts all use the capacity estimated from each fold's training rows; the full-record value only normalises the reported metrics. After
removing partial days at the beginning and end of the record, the dataset contains 391 complete UTC days.

\subsection{Weather Inputs}
\label{sec:weather-inputs}

Hourly weather data are obtained at the plant location through the Open-Meteo historical archive~\cite{zippenfenig2023openmeteo}. The product used for the main results is the operational analysis from the European Centre for Medium-Range Weather Forecasts Integrated Forecasting System (ECMWF IFS)~\cite{ecmwf2024ifs}. The provider reports this product at roughly $9$~km spatial resolution, and the grid point returned for the site is $50.93^{\circ}$N, $1.45^{\circ}$W. We request this product explicitly rather than relying on automatic product selection so that repeated runs use the same weather source. All weather values are returned at one-hour resolution in UTC and joined to
the corrected PV series by timestamp. Any remaining gaps in the weather inputs are filled using the nearest available value through forward or backward propagation. This filling is applied only to input variables; missing PV targets remain masked as described in \Cref{sec:site-data}. 
The weather vector $\mathbf{w}_\tau$ contains 15 variables:
\begin{align}
\mathbf{w}_\tau = (&
\text{shortwave radiation},\;
\text{direct normal irradiance},\;
\text{direct radiation},\;
\text{diffuse radiation},
\nonumber\\
&
\text{total cloud cover},\;
\text{low-level cloud cover},\;
\text{mid-level cloud cover},\;
\text{high-level cloud cover},
\nonumber\\
&
\text{2-metre temperature},\;
\text{relative humidity},\;
\text{dew-point temperature},
\nonumber\\
&
\text{10-metre wind speed},\;
\text{wind gust},\;
\text{precipitation},\;
\text{mean sea-level pressure}).
\end{align}

The analysis weather used for the main experiments is retrospective rather than a forecast issued before the target day. Results based on these inputs therefore cannot by themselves be interpreted as deployment accuracy.
\Cref{app:weather} repeats the pipeline using ERA5 reanalysis (ECMWF Re-Analysis version 5)~\cite{hersbach2020era5} and an archived forecast
product taken at a constant 24-hour lead, and measures the resulting change in forecast error.

\subsection{Solar Geometry and Temporal Features}
\label{sec:solar-features}

We compute solar position and irradiance quantities using
\texttt{pvlib}~\cite{holmgren2018pvlib} and the National Renewable Energy Laboratory Solar Position Algorithm~\cite{reda2004spa}. Ten derived features are supplied to the tabular models: solar zenith, solar elevation, cosine of the solar zenith, relative air mass, extraterrestrial direct normal irradiance, extraterrestrial irradiance on the horizontal plane, angle of incidence on the tilted array, cosine of the angle of incidence, POA global irradiance, and the clearness index $\mathrm{kt}_\tau$. Solar azimuth is also calculated because it is needed to
derive the angle of incidence and POA irradiance, but azimuth itself is not included in the model input. The clearness index is defined as $\mathrm{kt}_\tau
    =
    \frac{G_\tau}{G_\tau^{\mathrm{TOA}}},$
where $G_\tau$ is shortwave radiation and
$G_\tau^{\mathrm{TOA}}$ is extraterrestrial irradiance in the horizontal
plane. The index is set to zero outside daylight and clipped to $[0,1.5]$.
Dividing by extraterrestrial irradiance removes much of the deterministic solar irradiance scale and leaves a dimensionless measure of atmospheric attenuation. The upper limit is greater than one because the measured global horizontal irradiance can briefly exceed the corresponding extraterrestrial reference under cloud-enhancement conditions~\cite{vamvakas2020enhancement}.
In our evaluated data, this upper limit does not change any feature value: the highest clearness-index value across the 3869 held-out daylight hours is $1.004$. A separate bound applied to the POA-normalised prediction target is described in \Cref{sec:models}. 

Rather than adding categorical hour-of-day variables, we supply solar position features that vary with the position of the Sun, and seasonal position within the year is represented by
$
    \mathrm{doy}_{\sin}(\tau)=\sin\!\left(\tfrac{2\pi\,\mathrm{doy}(\tau)}{365.25}\right),
$
$
    \mathrm{doy}_{\cos}(\tau)=\cos\!\left(\tfrac{2\pi\,\mathrm{doy}(\tau)}{365.25}\right).
$
Short-term context is constructed for 11 quantities: shortwave radiation, direct normal irradiance, direct radiation, diffuse radiation, total, low-level, mid-level and high-level cloud cover, 10-metre wind speed, POA global irradiance, and the clearness index. For each quantity $z$, we add its one-hour lag, one-hour lead, and centred three-hour mean:
$
    z_{\tau-1},
$
$
    z_{\tau+1},
$
$
    \frac{z_{\tau-1}+z_\tau+z_{\tau+1}}{3}.
$
These operations produce $11\times3=33$ short-term context features. The final tabular input, $\boldsymbol{\chi}_\tau^{\mathrm{PV}}$, therefore contains 60 features: 15 weather variables, 10 solar geometry and clearness features, 33 short-term context features, and two day-of-year encodings. The convolutional model does not use the
60-dimensional vector directly. Instead, it receives weather information on the day-by-hour grid using six channels: shortwave radiation, direct normal irradiance, total cloud cover, 2-metre temperature, 2-metre relative humidity, and precipitation. The remaining inputs and their structure are defined in \Cref{sec:models}.

\section{Forecasting Models}
\label{sec:models}

The modelling stage has two parts. \Cref{sec:base-learners} describes five base learners that use different representations of the forecasting problem, while \Cref{sec:fusion} explains how their predictions are combined using weights learned only from validation data.

\subsection{Base Learners}
\label{sec:base-learners}
Ridge regression, a direct gradient boosting machine (GBM), a POA-normalised GBM, and an hour-specific GBM all use the tabular features defined in \Cref{sec:solar-features}; the CNN uses the day-by-hour representation described at the end of this subsection.

Ridge regression provides a linear reference model with $\ell_2$
regularisation, which is commonly used to control coefficient size when predictors are correlated~\cite{hoerl1970ridge}. Given the feature vector $\boldsymbol{\chi}_\tau^{\mathrm{PV}}$ defined in \Cref{sec:solar-features}, the prediction is $ \widehat{Y}_\tau^{\mathrm{PV}}
    =
    \max\left(
        0,
        \beta_0
        +
        \boldsymbol{\beta}^{\top}
        \boldsymbol{\chi}_\tau^{\mathrm{PV}}
    \right),$
where $\beta_0$ is the intercept and $\boldsymbol{\beta}$ is the coefficient vector. Features are standardised using training-set statistics only, and the coefficients are estimated by $\ell_2$-regularised least squares with a regularisation strength of 1. The final clipping operation enforces non-negative PV predictions. The main nonlinear tabular learner is a histogram-based GBM, following the gradient boosting framework of~\cite{friedman2001greedy} and implemented in scikit-learn~\cite{pedregosa2011scikit}. The model can represent nonlinear relationships and interactions among the features. Predictions are clipped to be non-negative.
We also train a POA-normalised GBM. Instead of predicting PV power directly, this model predicts
\begin{align}
    \kappa_\tau
    =
    \mathrm{clip}\left(
        \frac{Y_\tau^{\mathrm{PV}}}
        {\max\left(
            C^{\mathrm{PV}}
            \mathrm{POA}_\tau/1000,
            1
        \right)},
        0,
        1.5
    \right),
\end{align}
where $\mathrm{POA}_\tau$ is POA global irradiance and the capacity $C^{\mathrm{PV}}$ is estimated from the training portion of each fold as described in \Cref{sec:site-data}. The model is fitted only on daylight rows satisfying $\mathrm{POA}_\tau>50$~W\,m$^{-2}$. This fixed modelling threshold excludes very low-irradiance periods, where normalisation by the irradiance-based scale can produce unstable target values. Its predictions are converted back to PV power by multiplying by the same denominator, while non-daylight predictions are set to zero. This changes the prediction problem from estimating PV power directly to estimating output relative to an irradiance-based scale. The bound in this target serves a different purpose from the clearness-index feature in \Cref{sec:solar-features}. In the present data, the unbounded $\kappa_\tau$ exceeds $1.5$ on $10.3\%$ of daylight observations and reaches $21.7$. These large values occur when measured site-level PV generation is large relative to the irradiance-based denominator. The affected hours have
a median gridded cloud cover of $99\%$ while the plant is generating, which is consistent with disagreement between conditions represented by the weather grid cell and conditions at the PV site. We therefore clip the training target before squared error fitting. The value $1.5$ is a modelling
choice rather than a claimed physical limit. The sensitivity analysis in \Cref{app:clip} varies this bound from $1.0$ to no upper bound; across that range, the reported ensemble's daylight normalised root mean square error (nRMSE) changes by $0.03\%$ under random day-fold evaluation and by $0.29\%$ under rolling-origin evaluation.

The hour-specific GBM uses the same tabular features but fits a separate
prediction function, $f_h$, for each hour of the day:
$
f_h:
\boldsymbol{\chi}_\tau^{\mathrm{PV}}
\mapsto
\widehat{Y}_\tau^{\mathrm{PV}},
$
$h=0,\ldots,23.
$
Each $f_h$ is trained only on observations satisfying
$\mathrm{hour}(\tau)=h$. This allows the mapping from the input features to PV output to differ across hours rather than requiring one GBM to use the same mapping throughout the daily profile.

The fifth base learner is a two-dimensional CNN~\cite{lecun1998gradient}. Its input is a nine-channel $8\times24$ grid over the target day and the seven days before it: normalised PV history, an indicator marking the history days, the availability mask $a_\tau$ of \Cref{sec:site-data}, and the six weather channels. Target-day PV is set to zero. PV from validation and test days is also set to zero wherever it appears as history, and the corresponding availability entries are set to zero, so those days cannot enter another sample as past input. Under rolling-origin evaluation, this is a fixed-window forecast: the CNN is trained once and then predicts the whole test window without seeing new PV. Validation-day PV would be known at the origin; we still withhold it so validation is used only for stopping and stacking, a conservative choice. Missing target hours receive zero loss weight. Weather channels are standardised using training days only. Network predictions are scaled back to watts using the fold training capacity. The remaining settings are listed in \Cref{app:settings}. Unless otherwise stated, the reported CNN uses deterministic fold-specific training seeds with seed offset $0$; \Cref{sec:target-day-results} reports a five-run robustness check over offsets $0$--$4$, and \Cref{app:mask} uses the same seed offset as the headline run.

\subsection{Ensemble Fusion}
\label{sec:fusion}

Because the five base learners represent PV generation differently and make partly different errors, we test whether combining their predictions is more accurate than relying on a single model. Let
$\mathcal{M}_{\mathrm{PV}}=\{$Ridge, GBM, POA-normalised GBM, per-hour GBM, CNN$\}$
denote the learners, with prediction
$\widehat{Y}_{\tau,m}^{\mathrm{PV}}$ from learner $m$.
Within each evaluation fold, combination weights are fitted only on daylight, non-missing validation observations, so that generating hours determine the weights and the test data remain unseen. The fitted weights are then applied unchanged to the test predictions.
We compare four ways of combining the models. The arithmetic mean gives all learners equal weight, while inverse-validation-RMSE weighting gives more weight to models with lower validation error. Ridge stacking learns a regularised linear combination of the validation predictions. Our default is NNLS stacking~\cite{lawson1995solving}, which learns the combination directly from validation performance while preventing negative weights and cancellation between models. On the validation set $\mathcal{S}_{\mathrm{val}}$ it solves
\begin{align}
    \widehat{\boldsymbol{\alpha}}
    =
    \operatorname*{arg\,min}_{\boldsymbol{\alpha}\geq0}
    \sum_{\tau\in\mathcal{S}_{\mathrm{val}}}
    \Bigl(
        Y_\tau^{\mathrm{PV}}
        -
        \sum_{m\in\mathcal{M}_{\mathrm{PV}}}
        \alpha_m
        \widehat{Y}_{\tau,m}^{\mathrm{PV}}
    \Bigr)^2,
\end{align}
and predicts $\widehat{Y}_\tau^{\mathrm{NNLS}}=\sum_m\widehat{\alpha}_m\widehat{Y}_{\tau,m}^{\mathrm{PV}}$ using those coefficients unchanged, so the combination applied to the test rows is the minimiser of the objective above. The coefficients are not rescaled to sum to one, because dividing them by their sum would move the forecast away from that minimiser. Their sum is close to one in practice, averaging $1.03$ under random day-fold evaluation and $1.10$ under rolling-origin evaluation, so the fitted combination also applies a small rescaling of the base forecasts. In \Cref{fig:weights}, the weights are divided by their sum
for display as shares only; \Cref{sec:supp} reports what the two ways of forcing them to sum to one would cost. Non-negativity avoids cancellation between learners and makes the fitted combination easier to interpret.

\section{Evaluation}
\label{sec:evaluation}

This section defines the two evaluation protocols in \Cref{sec:protocols},
the accuracy measures and reference forecasts in \Cref{sec:metrics}, and the
main results with retrospective target-day weather in
\Cref{sec:target-day-results}. The appendices report the supporting diagnostics, the pipeline ablation, the sensitivity of the results to the weather input and the target bound, the computational cost of each learner, and the model settings.

\subsection{Evaluation Protocols}
\label{sec:protocols}

We evaluate every model under two complementary protocols. The rolling-origin protocol takes the first 120 complete days as the initial training pool and splits the remaining days into four contiguous test windows: 4 July--8 September 2025, 9 September--14 November 2025, 15 November 2025--20 January 2026, and 21 January--31 March 2026. For each window,
the models are fitted using only data preceding that window and are then evaluated on the window itself. Within each fold, the last $15\%$ of the unique training days, or seven days if that is larger, are held out as whole calendar days and used to fit the ensemble weights and to stop CNN training. The GBMs instead stop using an internal split of their training rows only. The first 120 days are used only for initial training and are not scored. The random day-fold protocol partitions complete days into five folds
uniformly at random using seed $0$. Each day appears in one test fold, so each observed target hour receives one held-out prediction across the five folds. Within each fold $k=0,\ldots,4$, either seven days or $10\%$ of the remaining days, whichever is larger, are sampled using seed $k$ as the validation set used for CNN early stopping and ensemble fitting.
Rolling-origin evaluation preserves time order; random day-fold evaluation mixes days across the record and has broader seasonal coverage. We report both.

\subsection{Metrics and Daylight Definition}
\label{sec:metrics}

Over a held-out set $\mathcal{S}_{\mathrm{test}}$ of predictions $\widehat{Y}_\tau^{\mathrm{PV}}$ with mean observed output $\bar{Y}^{\mathrm{PV}}$, we report
$
R^2=1-\frac{\sum_\tau(Y_\tau^{\mathrm{PV}}-\widehat{Y}_\tau^{\mathrm{PV}})^2}
{\sum_\tau(Y_\tau^{\mathrm{PV}}-\bar{Y}^{\mathrm{PV}})^2},$
the mean absolute error
$
\mathrm{MAE}=\frac{1}{N}\sum_\tau|Y_\tau^{\mathrm{PV}}-\widehat{Y}_\tau^{\mathrm{PV}}|,$
and
$
\mathrm{RMSE}=\sqrt{\frac{1}{N}\sum_\tau(Y_\tau^{\mathrm{PV}}-\widehat{Y}_\tau^{\mathrm{PV}})^2},
$
where $N=|\mathcal{S}_{\mathrm{test}}|$ and all sums are over $\mathcal{S}_{\mathrm{test}}$. The normalised metrics are
$\mathrm{nMAE}=100\,\mathrm{MAE}/C^{\mathrm{PV}}$ and
$\mathrm{nRMSE}=100\,\mathrm{RMSE}/C^{\mathrm{PV}}$, using the fixed reporting capacity $C^{\mathrm{PV}}\approx1664$~W defined in \Cref{sec:site-data}; relative changes quoted in the text are calculated from unrounded RMSE values.
Metrics are reported over all observed hours and over daylight hours, defined geometrically by solar elevation above $5^{\circ}$ using \texttt{pvlib} and the NREL Solar Position Algorithm~\cite{holmgren2018pvlib,reda2004spa}.
Because this definition depends only on solar position, it does not use held-out PV targets.

We compare the learned forecasts with three references~\cite{yang2020verification}: diurnal persistence uses the PV value 24 hours earlier, training-set climatology uses the mean output for each month and hour, and smart persistence carries the previous day's clearness ratio forward and rescales it using target-day POA irradiance,
$
\widehat{Y}_\tau^{\mathrm{SP}}
=
\frac{Y_{\tau-24}^{\mathrm{PV}}}
{\max(C^{\mathrm{PV}}_{\mathrm{tr}}\mathrm{POA}_{\tau-24}/1000,1)}
\,C^{\mathrm{PV}}_{\mathrm{tr}}\frac{\mathrm{POA}_\tau}{1000}.
$
The irradiance-based scale uses the weather product's POA irradiance rather than a clear-sky model, so this reference is an irradiance-scaled persistence that uses target-day weather through $\mathrm{POA}_\tau$. The carried ratio is bounded as in \Cref{sec:base-learners}, so the capacity does not cancel between the ratio and the rescaling; $C^{\mathrm{PV}}_{\mathrm{tr}}$ is therefore the fold's training capacity. Capacity and climatology fallbacks use that fold's training rows. Persistence uses the preceding observed day's PV when it is available at forecast time, including later days of a rolling test window; the CNN uses the stricter fixed-window history mask in \Cref{sec:base-learners}, so the comparison favours persistence in access to recent PV.

Where an input is unrecorded, a daylight zero would inflate the reference error, so both persistence references fall back to the last observation at the same hour within seven days and then to the training climatology. That window is exhausted for $1.1\%$ and $0.7\%$ of scored daylight hours under the two protocols. Climatology uses the training mean for the target month and hour, falling back to the mean for that hour across the months present. That fallback is needed for $53.8\%$ of rolling-origin daylight hours, where early folds meet months they have not seen, but the hour-of-day shape is always available, so the overall training mean is never used. Forecast skill is $s=1-\mathrm{RMSE}_{\mathrm{model}}/\mathrm{RMSE}_{\mathrm{reference}}$, where $s=0$ denotes equal RMSE and $s=1$ zero model error. We also evaluate the optimised climatology--persistence reference of~\cite{yang2019standard,yang2020verification} as a diagnostic: its weight is fitted leave-one-fold-out, so under rolling-origin evaluation a later fold can inform an earlier one and the combination is not a strictly chronological simulation. Smart persistence is the most accurate of these references under both protocols, at $18.07\%$ against $18.18\%$ daylight nRMSE under random day-fold evaluation and $18.63\%$ against $22.28\%$ under rolling-origin evaluation, so it is the reference for the reported skill scores. Restricting the comparison to the $98.2\%$ and $98.8\%$ of hours whose $24$-hour lag was recorded, where no reference falls back at all, moves the skill of the reported ensemble to $30.5\%$ and $2.2\%$, so the reported skill does not rest on hours where an input had to be replaced.

\subsection{Results with Target-Day Weather}
\label{sec:target-day-results}

\begin{table}[t]
\centering
\caption{Selected PV forecasting results using retrospective target-day weather.}
\label{tab:forecasting-results-selected}
\small
\begin{adjustbox}{max width=\linewidth}
\begin{tabular}{llrrrrrrr}
\toprule
\textbf{Protocol} & \textbf{Model} & \textbf{Subset}
& $n$ & $R^2$ & MAE & RMSE & nMAE & nRMSE \\
& & & & & (W) & (W) & (\%) & (\%) \\
\midrule
Rolling-origin & NNLS stacking      & All      & 6076 & 0.704 &
86.86 & 194.35 & 5.22 & 11.68 \\

Rolling-origin & NNLS stacking      & Daylight & 2517 & 0.506 &
199.32 & 301.03 & 11.98 & 18.09 \\

Rolling-origin & Ridge stacking     & Daylight & 2517 & 0.546 &
192.04 & 288.52 & 11.54 & 17.34 \\

Rolling-origin & Validation-selected & Daylight & 2517 & 0.474 &
204.31 & 310.43 & 12.28 & 18.65 \\

Rolling-origin & Smart persistence  & Daylight & 2517 & 0.476 &
203.46 & 309.97 & 12.23 & 18.63 \\

Random day-fold & NNLS stacking     & All      & 8516 & 0.859 &
69.00 & 139.88 & 4.15 & 8.41 \\

Random day-fold & NNLS stacking     & Daylight & 3869 & 0.765 &
141.03 & 206.76 & 8.47 & 12.42 \\

Random day-fold & Ridge stacking    & Daylight & 3869 & 0.761 &
143.35 & 208.32 & 8.61 & 12.52 \\

Random day-fold & Validation-selected & Daylight & 3869 & 0.730 &
154.54 & 221.50 & 9.29 & 13.31 \\

Random day-fold & Smart persistence & Daylight & 3869 & 0.503 &
196.14 & 300.73 & 11.79 & 18.07 \\
\bottomrule
\end{tabular}
\end{adjustbox}
\end{table}

\Cref{tab:forecasting-results-selected} shows a substantial difference between the two protocols\footnote{$n$ is the number of non-missing hourly observations included in each reported subset.}. For NNLS stacking, daylight $R^2$ falls from $0.765$ under random day-fold evaluation to $0.506$ under rolling-origin evaluation, and daylight nRMSE rises from $12.42\%$ to $18.09\%$ of capacity. The protocols also differ in seasonal coverage, so the gap cannot be assigned to time order alone, but it shows that performance is lower when later periods must be predicted from earlier observations alone. Daylight results are also weaker than all-hours results: under random day-fold evaluation, $R^2$ decreases from $0.859$ over all observed hours to $0.765$ over daylight hours, because the all-hours subset contains night-time periods for which PV output is normally zero. Both subsets are retained because the pipeline produces complete 24-hour profiles while the daylight subset shows performance during generating hours.

NNLS stacking is the pre-specified combination. Its comparator is the validation-selected base learner: in each fold, the member with the lowest daylight validation RMSE, so the reported row can use different learners in different folds, and neither side is chosen on the test data. Under random day-fold evaluation the stack reaches daylight $R^2=0.765$ and RMSE $=206.76$~W against $R^2=0.730$ and RMSE $=221.50$~W for that comparator, a $6.7\%$ reduction in RMSE. Under rolling-origin evaluation, it reduces daylight RMSE from $310.43$~W to $301.03$~W, a $3.0\%$ reduction. Versus smart persistence, the reduction is $2.9\%$, with a $95\%$ day-level bootstrap interval $[-6.3,11.4]\%$ that includes zero, so we do not treat it as day-robust. Hourly errors within a day are dependent, so we do not use an hourly equal-accuracy test~\cite{diebold1995comparing}. Conditional on the fitted out-of-fold forecasts, a paired block bootstrap~\cite{kunsch1989jackknife} resamples whole target days over $10^4$ replicates with random number generator (RNG) seed $0$. For the NNLS-versus-validation-selected reductions, the $95\%$ intervals are $[4.1,9.2]\%$ over the $358$ random-fold days with a scored daylight hour and $[1.5,4.4]\%$ over the $254$ such rolling-origin days. For random day-fold evaluation, 3-day and 7-day blocks give $[3.8,9.6]\%$ and $[3.0,10.6]\%$; for rolling-origin evaluation they give $[1.0,4.8]\%$ and $[0.4,5.1]\%$. A five-run check over offsets $0$--$4$ varies the CNN seed offset and the random day partition, not the rolling windows, giving daylight nRMSE $12.48\pm0.14\%$ and $18.07\pm0.12\%$ against $13.32\pm0.18\%$ and $18.78\pm0.28\%$ for the comparator.

Ridge stacking reaches a lower rolling-origin daylight nRMSE, $17.34\%$ against $18.09\%$. Identifying it as the stronger method requires the test metrics, so we record it as an observation rather than as the accuracy of the pipeline, and every performance comparison uses NNLS stacking as the designated ensemble. Fitting all five learners takes $12.9$~s against $1.0$~s for the direct GBM, and a forecast day costs $0.7$~ms against $0.1$~ms (\Cref{app:cost}). On the measured hardware this is small relative to the 24-hour forecast horizon; where it is not, dropping to the validation-selected single model gives up the reductions above. \Cref{tab:forecasting-results-selected} also shows that no single learner is best under both protocols, whereas the combination is close to the best under each.
\Cref{sec:ablation} shows that the $-1$\,h correction costs $0.40$ and $1.01$ percentage points of daylight nRMSE, and that dropping the raw solar and clearness inputs while keeping the POA-normalised target lowers rolling-origin daylight nRMSE from $18.09\%$ to $15.00\%$. \Cref{app:weather} shows that the constant-lead forecast product raises daylight RMSE by $13.1\%$ and $4.2\%$ relative to the matched analysis control.
\section{Conclusion and Future Work}
\label{sec:conclusion}

This paper presented a day-ahead PV forecasting pipeline for a site with about one year of measurements and public weather inputs. NNLS stacking reaches daylight $R^2=0.765$ under random day-fold evaluation and $0.506$ under rolling-origin evaluation, falling to $0.459$ in a sensitivity analysis that replaces the retrospective weather with a public forecast product taken at a constant 24-hour lead. The ablation shows that the timestamp correction mainly helps the physical alignment check, that short-term weather context helps rolling-origin performance, and that removing the raw solar and clearness inputs while keeping the POA-normalised target substantially improves accuracy on the short rolling histories studied here. These findings may not transfer to other sites, climates, or longer records; array geometry is assumed; and the forecast-input comparison uses a constant-lead product rather than a replay of one issuance, so it is not the penalty of a single weather model. The commitment-cost check in \Cref{app:opsvalue} shows that a model selected on squared error need not be the one that commits best under an asymmetric penalty. Future work should therefore test the pipeline on more sites and longer records, and replace the single-valued forecasts with probabilistic ones that can be used directly in charging and storage decisions.

\section*{Acknowledgements}
\addcontentsline{toc}{section}{Acknowledgements}

This work was supported by EPSRC through the Turing AI Fellowship “Citizen-Centric AI Systems” (EP/V022067/1) and the FEVER project (EP/W005883/1). We also thank the IRIDIS High Performance Computing Facility and the Low Carbon Comfort Centre at the University of Southampton.

\clearpage
\bibliographystyle{ieeetr}
\bibliography{references}


\clearpage

\appendix
\crefalias{section}{appendix}

\section{Forecast Quality and Error Diversity}
\label{sec:supp}

\begin{figure}[!ht]
\centering
\includegraphics[width=0.65\linewidth]{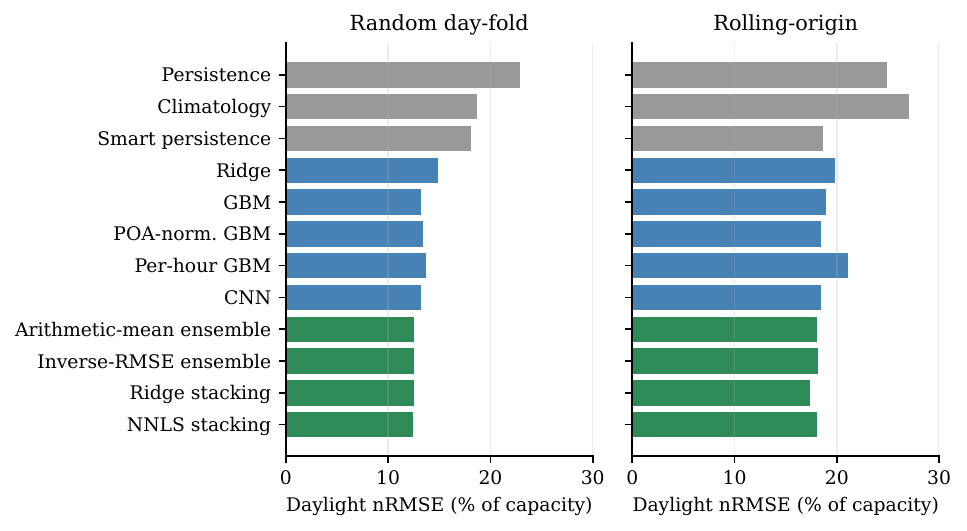}
\caption{Daylight nRMSE of every model and the three main reference forecasts under both evaluation protocols.}
\label{fig:skill-baselines}
\end{figure}

\begin{figure}[!ht]
\centering
\includegraphics[width=0.65\linewidth]{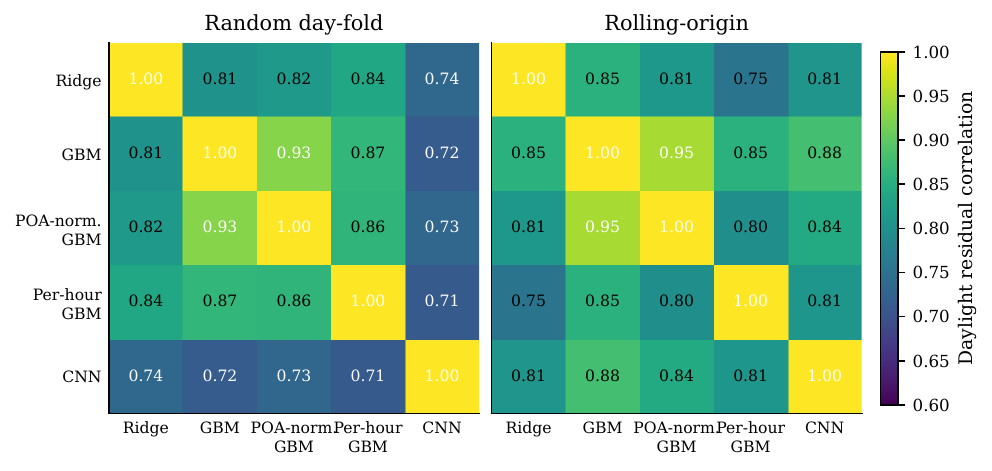}
\caption{Pairwise correlation of the five base-learner residuals on daylight hours under random day-fold (left) and rolling-origin (right) evaluation.}
\label{fig:residual-corr}
\end{figure}

\begin{figure}[!ht]
\centering
\includegraphics[width=0.65\linewidth]{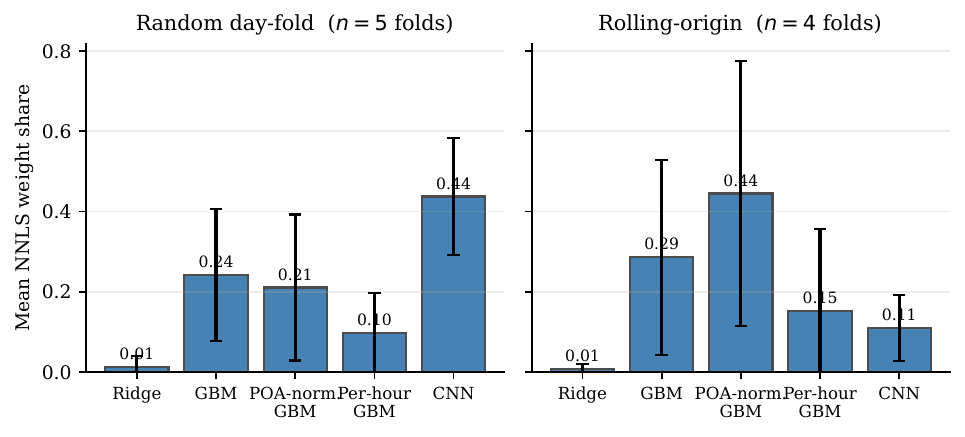}
\caption{Mean ($\pm$ standard deviation across folds) NNLS stacking weight for each base learner, divided by the fold's weight sum so that the bars read as shares.}
\label{fig:weights}
\end{figure}

\Cref{fig:skill-baselines} places every learner and the three main references on the same daylight nRMSE scale\footnote{``POA-norm.'' abbreviates ``POA-normalised.''}. In \Cref{fig:residual-corr}, the CNN has the lowest residual correlations with the other learners under random day-fold evaluation, approximately $0.71$--$0.74$, rising to $0.81$--$0.88$ under rolling-origin evaluation; those values show that the errors are less similar under random folds, not that residual diversity causes the ensemble gain. In \Cref{fig:weights}, the CNN receives the largest mean NNLS weight under random day-fold evaluation, approximately $0.44$, while under rolling-origin evaluation the POA-normalised GBM receives approximately $0.44$ and the CNN weight falls to approximately $0.11$; Ridge receives little or no mean weight under either protocol. The weights show how the fitted combination changes; they are not a measure of importance. Rescaling the fitted coefficients to sum to one gives daylight nRMSE of $12.30\%$ and $18.71\%$, against $12.42\%$ and $18.09\%$ as fitted, and imposing the constraint during fitting gives $12.54\%$ and $18.10\%$. We keep the coefficients as fitted, which is the only one of the three that solves the problem stated in \Cref{sec:fusion}.

\section{Pipeline Ablation}
\label{sec:ablation}

\begin{figure}[!ht]
\centering
\includegraphics[width=0.55\linewidth]{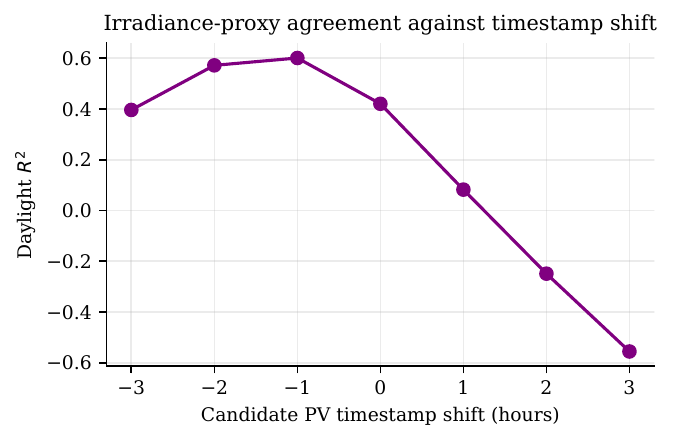}
\caption{Alignment diagnostic: daylight $R^2$ of the irradiance check with no fitted parameters as a function of the candidate PV timestamp shift.}
\label{fig:alignment}
\end{figure}

This appendix tests three parts of the pipeline: timestamp-alignment (\Cref{fig:alignment}), the solar and clearness input features, and the short-term temporal context. All were run after the main test results were obtained, so they are diagnostics rather than grounds for changing the reported setup.
The $-1$\,h PV shift was chosen on the full record by sweeping candidate shifts against an irradiance check with no fitted parameters,
$
    \widehat{Y}_\tau
    =
    C^{\mathrm{PV}}
    \mathrm{GHI}_\tau/1000
$.
Applying this correction increases the daylight $R^2$ of that check from $0.42$ to $0.60$ and moves the peak irradiance--power correlation to zero lag.
Sweeping the full record uses hours that later become test data, so we repeated the sweep on the training days of each fold alone, taking the capacity from those days as well. That training-only sweep selects $-1$\,h in seven of the nine fold training sets, including all five random day-folds. The two exceptions are the first and third rolling-origin windows, which prefer $-2$\,h by $0.004$ in $R^2$; the same two candidates rank first and second everywhere else. The correction is therefore recoverable from training rows alone, although in two of the four rolling-origin folds those rows would have selected $-2$\,h rather than the $-1$\,h applied throughout. We report the single global correction because it is one stated timestamp convention rather than a per-fold one. The two candidates differ by only $0.004$ in $R^2$ in those folds.
The first three rows of \Cref{tab:ablation} show that the learned models are slightly more accurate with no correction, under both protocols, with the $-2$\,h candidate in between: the irradiance check has no free parameters, so a one-hour offset passes straight into its error, whereas the learners receive solar geometry, calendar, and lagged weather features from which a constant offset can largely be absorbed. We keep the $-1$\,h correction as the stated timestamp convention supported by the physical check, not because it minimises learned-model error. No shift is more accurate for the learners ($12.02\%$ against $12.42\%$ daylight nRMSE under random day-fold evaluation and $17.08\%$ against $18.09\%$ under rolling-origin evaluation), and we report that cost rather than putting the no-shift pipeline in the headline. The cost of doing so is $0.40$ percentage points of daylight nRMSE under random day-fold evaluation and $1.01$ under rolling-origin evaluation. The check does not identify the cause: British Summer Time hours give their largest $R^2$ at $-2$\,h ($0.657$, versus $0.653$ at $-1$\,h) and Greenwich Mean Time hours at $0$\,h ($0.369$, versus $0.363$ at $-1$\,h), and a daylight-saving-aware correction changes the whole-record $R^2$ only from $0.601$ to $0.603$.
The other two ablations act mainly on the rolling-origin protocol. Removing the lag, lead, and rolling features leaves random day-fold error almost unchanged, $12.42\%$ to $12.49\%$, but raises rolling-origin nRMSE from $18.09\%$ to $18.19\%$. Removing the raw solar and clearness input features while keeping the POA-normalised target again barely moves the random day-fold result, yet improves rolling-origin daylight nRMSE from $18.09\%$ to $15.00\%$ ($R^2$ from $0.506$ to $0.660$). Those inputs vary strongly with season, and the early rolling-origin folds have seen little of the year, so poor extrapolation is one possible reading; the ablation does not isolate that cause.

\begin{table}[!ht]
\centering
\caption{Pipeline ablation on the daylight subset, reporting NNLS stacking in every row.}
\label{tab:ablation}
\small
\begin{adjustbox}{max width=\linewidth}
\begin{tabular}{lrrrr}
\toprule
& \multicolumn{2}{c}{\textbf{Random day-fold}} &
\multicolumn{2}{c}{\textbf{Rolling-origin}} \\
\cmidrule(lr){2-3}\cmidrule(lr){4-5}
\textbf{Configuration} & $R^2$ & nRMSE (\%) &
$R^2$ & nRMSE (\%) \\
\midrule
Full pipeline ($-1$\,h shift)       & 0.765 & 12.42 & 0.506 & 18.09 \\
$-2$\,h alignment shift             & 0.764 & 12.55 & 0.532 & 17.81 \\
No alignment shift                  & 0.788 & 12.02 & 0.577 & 17.08 \\
No solar/clearness inputs           & 0.763 & 12.46 & 0.660 & 15.00 \\
No temporal context                 & 0.763 & 12.49 & 0.500 & 18.19 \\
\bottomrule
\end{tabular}
\end{adjustbox}
\end{table}

\section{Choice of Weather Product}
\label{app:weather}
The forecast inputs come from the Open-Meteo previous-runs archive in UTC. For each valid hour, this archive returns the value predicted 24 hours earlier, so each 24-hour profile is assembled from a sequence of model updates at a constant 24-hour lead rather than being read from a single forecast run with a single issue time and a single data cutoff. The request does not name a model, so the product is not pinned to the ECMWF IFS. Twelve of the 15 analysis variables are available. Because the one-hour lead and the centred three-hour mean reach one hour past the end of the target day, the last hour of the profile draws on a valid time that falls on the following calendar day, and at a 24-hour lead that value originates on the target day itself. The run is therefore a sensitivity analysis of the weather inputs at a fixed 24-hour lead: the inputs are not those of any single issuance, and the gap from the analysis control combines forecast lead with a change of weather product.
\Cref{tab:weather-full} reports the matched 12-variable IFS-analysis control, ERA5, and the reference forecasts. Restricting the analysis input from 15 to 12 variables changes RMSE by $0.05\%$ under random day-fold evaluation and by $0.46\%$ under rolling-origin evaluation. The increases quoted in the main text are therefore relative to that matched control, not to the 15-variable headline run. Over daylight hours, the 24-hour-lead shortwave radiation differs from the analysis by $119$~W\,m$^{-2}$ RMSE (analysis mean $305$~W\,m$^{-2}$), and cloud cover correlates at $0.62$. ERA5 is worse than the IFS analysis under random day-fold evaluation and better under rolling-origin evaluation; with one site we cannot identify the reason. Persistence and hourly climatology do not change with the weather source, as expected because they use no weather. Smart persistence does change, because it uses target-day POA irradiance.

\begin{table}[!ht]
\centering
\caption{Daylight results by weather input for NNLS stacking, together with the reference forecasts.}
\label{tab:weather-full}
\small
\begin{adjustbox}{max width=\linewidth}
\begin{tabular}{llrrrrr}
\toprule
& & & \multicolumn{4}{c}{\textbf{nRMSE (\%) of capacity}} \\
\cmidrule(l){4-7}
\textbf{Protocol} & \textbf{Weather input} & $R^2$
& Ensemble & Smart pers. & Persistence & Climatology \\
\midrule
Random day-fold & IFS analysis, 15 variables
& 0.765 & 12.42 & 18.07 & 22.89 & 18.64 \\

Random day-fold & IFS analysis, 12 variables
& 0.765 & 12.43 & 18.07 & 22.89 & 18.64 \\

Random day-fold & ERA5 reanalysis, 15 variables
& 0.742 & 13.03 & 17.93 & 22.89 & 18.64 \\

Random day-fold & 24-hour-lead forecast product, 12 variables
& 0.699 & 14.05 & 21.04 & 22.89 & 18.64 \\

\midrule

Rolling-origin & IFS analysis, 15 variables
& 0.506 & 18.09 & 18.63 & 24.92 & 27.05 \\

Rolling-origin & IFS analysis, 12 variables
& 0.501 & 18.17 & 18.63 & 24.92 & 27.05 \\

Rolling-origin & ERA5 reanalysis, 15 variables
& 0.573 & 16.82 & 18.29 & 24.92 & 27.05 \\

Rolling-origin & 24-hour-lead forecast product, 12 variables
& 0.459 & 18.93 & 20.38 & 24.92 & 27.05 \\
\bottomrule
\end{tabular}
\end{adjustbox}
\end{table}

\section{Sensitivity to the POA-Normalised Target Bound}
\label{app:clip}
\Cref{tab:clip} varies only the POA-normalised target bound of \Cref{sec:base-learners}; the clearness-index feature bound is unchanged. Across the six settings, the reported ensemble's daylight nRMSE stays between $12.40\%$ and $12.43\%$ under random day-fold evaluation and between $17.80\%$ and $18.09\%$ under rolling-origin evaluation, so the result does not depend on this choice at the precision we report. The value $1.5$ is not uniquely favoured: the lowest rolling-origin error in the sweep is at $1.0$. We keep $1.5$ because it is the value the reported runs used, not one selected after seeing this sweep.

\begin{table}[!ht]
\centering
\caption{Effect of the POA-normalised target bound on the daylight subset.}
\label{tab:clip}
\small
\begin{adjustbox}{max width=\linewidth}
\begin{tabular}{lrrrr}
\toprule
& \multicolumn{2}{c}{\textbf{Random day-fold}} &
\multicolumn{2}{c}{\textbf{Rolling-origin}} \\
\cmidrule(lr){2-3}\cmidrule(lr){4-5}
\textbf{Bound} & POA-norm.\ GBM & Ensemble &
POA-norm.\ GBM & Ensemble \\
& nRMSE (\%) & nRMSE (\%) & nRMSE (\%) & nRMSE (\%) \\
\midrule
$1.0$        & 13.84 & 12.40 & 18.53 & 17.80 \\
$1.25$       & 13.41 & 12.43 & 18.40 & 17.97 \\
$1.5$ (used) & 13.38 & 12.42 & 18.43 & 18.09 \\
$2.0$        & 13.33 & 12.41 & 18.45 & 18.04 \\
$3.0$        & 13.37 & 12.41 & 18.26 & 17.83 \\
Unbounded    & 13.38 & 12.43 & 18.13 & 17.84 \\
\bottomrule
\end{tabular}
\end{adjustbox}
\end{table}

\section{Pricing Forecast Error as a Commitment Cost}
\label{app:opsvalue}

This appendix tests whether ranking models by an asymmetric hourly commitment penalty gives the same ordering as ranking them by forecast error.
For hourly commitment $c_\tau$ and realised generation
$Y_\tau^{\mathrm{PV}}$, we define
$
    L_r(c_\tau,Y_\tau^{\mathrm{PV}})
    =
    r\,(c_\tau-Y_\tau^{\mathrm{PV}})_+
    +
    (Y_\tau^{\mathrm{PV}}-c_\tau)_+,
$
where the coefficient on under-commitment, $(Y_\tau^{\mathrm{PV}}-c_\tau)_+$, is normalised to one, while $r$ is the relative penalty on over-commitment, $(c_\tau-Y_\tau^{\mathrm{PV}})_+$. Commitments are clipped at zero. We report the summed penalty per unit of delivered energy. This is a simplified decision proxy rather than a model of a particular market or charging station settlement.
This diagnostic uses the main retrospective-IFS forecast outputs and therefore should not be interpreted as an operational day-ahead settlement-cost estimate. The cost analysis uses all observed held-out hours because the commitment is a complete 24-hour profile. At $r=1$ the loss reduces to absolute error; as $r$ increases, over-commitment receives a larger penalty, so the ranking need not match an RMSE ranking. \Cref{tab:opsvalue} confirms this: at $r=3$, the POA-normalised GBM has a lower commitment cost than the reported NNLS stack under both protocols, despite the higher RMSE.
For a predictive distribution with cumulative distribution function $F$, minimising the expected value of the loss above gives
$
    F(c^*) = \frac{1}{r+1}.
$
The cost-optimal commitment is therefore a quantile rather than the mean, so a model selected on squared error need not be the one that commits best. This is why we treat probabilistic forecasting as the natural next step. The proxy settles every hour independently and omits storage, charging demand, and other station constraints, so it measures one consequence of forecast error under an asymmetric loss.

\begin{table}[!ht]
\centering
\caption{Day-ahead commitment cost per unit of delivered energy, evaluated
over all observed held-out hours.}
\label{tab:opsvalue}
\small
\begin{adjustbox}{max width=\linewidth}
\begin{tabular}{lrrrr}
\toprule
& & \multicolumn{2}{c}{\textbf{Commitment cost}} & \\
\cmidrule(lr){3-4}
\textbf{Model} & RMSE (W) & $r=1$ & $r=3$ &
\textbf{Saving at $r=3$} \\
\midrule
\multicolumn{5}{l}{\emph{Random day-fold}} \\

Persistence            & 256.9 & 0.537 & 1.054 & $-0.27$ \\
Hourly climatology     & 209.3 & 0.484 & 0.965 & $-0.17$ \\
Smart persistence      & 202.8 & 0.413 & 0.827 & $0.00$ \\
Ridge regression       & 167.6 & 0.408 & 0.844 & $-0.02$ \\
Gradient boosting      & 148.6 & \textbf{0.309} & 0.618 & $0.25$ \\
POA-normalised GBM     & 150.4 & 0.314 & \textbf{0.587} & $\mathbf{0.29}$ \\
Per-hour GBM           & 154.0 & 0.330 & 0.664 & $0.20$ \\
CNN                    & 150.2 & 0.364 & 0.749 & $0.09$ \\
Ridge stacking         & 143.1 & 0.355 & 0.782 & $0.05$ \\
NNLS stacking          & \textbf{139.9} & 0.316 & 0.668 & $0.19$ \\

\midrule
\multicolumn{5}{l}{\emph{Rolling-origin}} \\

Persistence            & 267.2 & 0.618 & 1.220 & $-0.39$ \\
Hourly climatology     & 311.3 & 0.942 & 2.214 & $-1.52$ \\
Smart persistence      & 199.7 & 0.440 & 0.879 & $0.00$ \\
Ridge regression       & 225.6 & 0.703 & 1.431 & $-0.63$ \\
Gradient boosting      & 203.7 & 0.474 & 0.786 & $0.11$ \\
POA-normalised GBM     & 197.9 & \textbf{0.440} & \textbf{0.653} & $\mathbf{0.26}$ \\
Per-hour GBM           & 229.7 & 0.575 & 0.987 & $-0.12$ \\
CNN                    & 211.2 & 0.651 & 1.316 & $-0.50$ \\
Ridge stacking         & \textbf{187.2} & 0.458 & 0.767 & $0.13$ \\
NNLS stacking          & 194.3 & 0.448 & 0.752 & $0.14$ \\
\bottomrule
\end{tabular}
\end{adjustbox}
\end{table}

\section{Computational Cost and Value of Each Base Learner}
\label{app:cost}

This appendix reports the measured fitting and prediction times in \Cref{tab:cost} and tests how NNLS ensemble accuracy changes when each base learner is removed.
The times are indicative implementation measurements on an AMD EPYC 9334 CPU with the thread settings used by the implementation (up to four CPU threads; no GPU), not hardware-independent benchmarks. Minimum software versions are Python~$\geq 3.10$, scikit-learn~$\geq 1.3$, and PyTorch~$\geq 2.0$, as listed in the repository requirements file.
The leave-one-learner-out results are protocol dependent. Under random day-fold evaluation, removing the CNN increases ensemble RMSE by $5.77\%$, while removing any other learner changes RMSE by less than $1\%$. Under rolling-origin evaluation, removing the POA-normalised GBM increases RMSE by $2.53\%$, whereas removing the CNN increases it by $0.46\%$. Removing Ridge does not increase RMSE in either protocol. This is consistent with the small NNLS stacking weights shown in \Cref{fig:weights}.
These omission experiments refit only the NNLS combination weights using saved fold predictions and leave-one-fold-out fitting. This is a diagnostic comparison and is not a strict rolling-origin simulation. Their values are therefore suitable for comparing omission cases with each other, but they are not numerically identical to the headline NNLS results, whose weights are fitted using the within-fold validation procedure described in \Cref{sec:fusion}.

\begin{table}[!ht]
\centering
\caption{Measured fitting and prediction cost of each base learner and the
change in NNLS RMSE when that learner is omitted.}
\label{tab:cost}
\small
\begin{adjustbox}{max width=\linewidth}
\begin{tabular}{lrrrr}
\toprule
& & & \multicolumn{2}{c}{\textbf{RMSE increase if dropped (\%)}} \\
\cmidrule(l){4-5}
\textbf{Base learner} & Fit (s) & Predict (ms/day) &
Random day-fold & Rolling-origin \\
\midrule
Ridge regression
& 0.01 & $<0.01$ & 0.00 & $-0.85$ \\

Gradient boosting
& 1.01 & 0.12 & 0.60 & $0.00$ \\

POA-normalised GBM
& 1.87 & 0.25 & 0.00 & 2.53 \\

Per-hour GBM
& 3.37 & 0.36 & 0.19 & $-0.07$ \\

CNN
& 6.59 & $<0.01$ & 5.77 & $0.46$ \\

\midrule
All five
& 12.85 & 0.73 & --- & --- \\
\bottomrule
\end{tabular}
\end{adjustbox}
\end{table}

\section{Missing Data Handling in the Convolutional Network}
\label{app:mask}

This appendix tests the two CNN missing data mechanisms introduced in \Cref{sec:site-data,sec:base-learners}: the availability input channel and zero loss weight for unobserved targets. The four combinations are compared in \Cref{tab:mask}.
When an unavailable historical PV value is filled with zero, the numerical value alone cannot distinguish that hour from a genuine zero-generation measurement. The full CNN therefore uses an availability channel $a_\tau=\mathbf{1}\{\tau\notin\mathcal{M}\}$ to mark observed historical hours. Separately, missing target hours receive zero loss weight and therefore do not contribute to the training gradient. \Cref{tab:mask} removes these two mechanisms individually and together. Each configuration uses the same CNN seed offset as the headline run.
The full configuration gives the lowest CNN nRMSE under random day-fold evaluation, $13.23\%$ against $13.44\%$ for the naive zero fill, and each mechanism used alone is worse than using neither, suggesting that the two mechanisms interact rather than contributing independently. Under rolling-origin evaluation, the four variants lie within $0.29$ percentage points and the ordering is not stable.
The ensemble tells a different story. Its daylight nRMSE spans $12.31\%$ to $12.46\%$ under random day-fold evaluation and $17.69\%$ to $18.09\%$ under rolling-origin evaluation, and the naive zero fill gives the lowest ensemble error only under random day-fold evaluation. The other four learners do not share the CNN's inputs, so they absorb the difference and can also gain from a slightly less correlated CNN. We keep the availability channel and the loss weighting because the full CNN performs best under random day-fold evaluation and because the two mechanisms explicitly distinguish unobserved hours from genuine zero generation; their rolling-origin differences are small and unstable. We do not keep them because they improve the reported ensemble. Missingness is also strongly clustered in this dataset: $91.9\%$ of missing hours occur in four outages longer than three days. The remaining missing hours are three isolated one-hour gaps and two shorter outages of $57$ and $10$ hours. These experiments support the semantic rationale for representing missingness explicitly in the CNN, but they do not show an ensemble-accuracy advantage for that choice, and they do not establish how the result would change for frequent short gaps.

\begin{table}[!ht]
\centering
\caption{Missing data handling in the CNN on the daylight subset, using the
same CNN seed offset as the headline run, with the resulting NNLS ensemble.}
\label{tab:mask}
\small
\begin{adjustbox}{max width=\linewidth}
\begin{tabular}{lrrrrrr}
\toprule
& \multicolumn{3}{c}{\textbf{Random day-fold}} &
\multicolumn{3}{c}{\textbf{Rolling-origin}} \\
\cmidrule(lr){2-4}\cmidrule(lr){5-7}
& \multicolumn{2}{c}{CNN} & Ensemble &
\multicolumn{2}{c}{CNN} & Ensemble \\
\cmidrule(lr){2-3}\cmidrule(lr){5-6}
\textbf{Configuration} & $R^2$ & nRMSE (\%) & nRMSE (\%) &
$R^2$ & nRMSE (\%) & nRMSE (\%) \\
\midrule
Channel and loss weight (used)
& 0.733 & \textbf{13.23} & 12.42 & 0.487 & 18.43 & 18.09 \\

Loss weight only
& 0.721 & 13.52 & 12.46 & 0.489 & 18.39 & 17.69 \\

Channel only
& 0.718 & 13.61 & 12.46 & 0.473 & 18.68 & 18.08 \\

Neither (naive fill with zero)
& 0.725 & 13.44 & \textbf{12.31} & 0.488 & 18.41 & 17.76 \\
\bottomrule
\end{tabular}
\end{adjustbox}
\end{table}

\section{Model Settings}
\label{app:settings}

\Cref{tab:settings} lists the settings not fully specified in \Cref{sec:models}, as they appear in the code that produced the reported results. They were fixed before any held-out test metric was inspected and then left unchanged: no search over them was run, and none is tuned per fold, protocol, or weather product. The only quantities estimated inside a fold are the model parameters, the ensemble weights, and the stopping point, each from training or validation rows.

\begin{table}[H]
\centering
\caption{Fixed model settings used in the reported experiments.}
\label{tab:settings}
\footnotesize
\begin{adjustbox}{max width=\linewidth}
\begin{tabular}{p{0.18\linewidth}p{0.36\linewidth}p{0.38\linewidth}}
\toprule
\textbf{Model} & \textbf{Model structure} & \textbf{Training and stopping} \\
\midrule

GBM and POA-normalised GBM
& Maximum depth 8; at least 30 samples per leaf; $\ell_2$ regularisation $0.1$.
& Up to 800 iterations; learning rate $0.04$; early stopping using an internal $12\%$ training split after 25 rounds without improvement; predictions clipped at zero. \\

Per-hour GBM
& Maximum depth 6; at least 15 samples per leaf; $\ell_2$ regularisation $0.2$.
& Up to 500 iterations; learning rate $0.05$; early stopping using an internal $15\%$ training split after 15 rounds without improvement. \\

Ridge stacking
& Linear combination with an intercept and regularisation strength $1$.
& Fitted to the validation predictions; final predictions clipped at zero. \\

CNN
& Three convolutional layers with kernel sizes $3{\times}5$, $3{\times}5$, and $3{\times}3$, using 32, 32, and 16 channels; GroupNorm~\cite{wu2018group}, GELU~\cite{hendrycks2016gaussian}, and dropout $0.1$.
& AdamW~\cite{loshchilov2019decoupled}; learning rate $2{\times}10^{-3}$; weight decay $10^{-4}$; batch size 32; at most 120 epochs; cosine learning-rate schedule; early stopping after 15 validation epochs without improvement. \\

\bottomrule
\end{tabular}
\end{adjustbox}
\end{table}

\end{document}